\pdfoutput=1

\documentclass[11pt,a4paper]{article}
\usepackage{amsmath}

\usepackage[acceptedWithA]{tacl2021v1}
\usepackage{times}
\usepackage{latexsym}

\usepackage[T1]{fontenc}

\usepackage[utf8]{inputenc}

\usepackage{microtype}

\usepackage{inconsolata}

\usepackage{graphicx}

\usepackage{booktabs}
\usepackage{CJKutf8}
\usepackage{array}
\usepackage{adjustbox}
\usepackage{url}

\usepackage{listings}
\usepackage{xcolor}
\usepackage[inline]{enumitem}
\usepackage{multirow}

\newcommand{\blue}[1]{\textcolor{black}{#1}}
\newcommand{\orange}[1]{\textcolor{black}{#1}}

\renewcommand{\confidential}{\textcolor{black}{\confidentialtext}}
\let\paperAtPageLowishCenter\AtPageLowishCenter
\renewcommand{\AtPageLowishCenter}[1]{\paperAtPageLowishCenter{\textcolor{black}{#1}}}

\title{A Systematic Analysis of \\the Predictive Power of LM Surprisal in Reading Chinese}

\author{
Hongao Zhu$^{\spadesuit*}$, Muxiaoqiao Xu$^{\dagger*}$, Yikang Liu$^\sharp$,  Siyuan Song$^{\dagger\diamond\spadesuit}$, \\
{\bf Yuxia Wang$^\dagger$, Byung-Doh Oh$^\ddagger$, Hai Hu$^\heartsuit$} \vspace{.3em}
\\
{\small $^\spadesuit$Department of Linguistics, UC San Diego \quad $^\dagger$School of Foreign Languages, Shanghai Jiao Tong University} \\
{\small $^\sharp$School of Computer Science, Shanghai Jiao Tong University \quad $^\diamond$Department of Psychology, Princeton University} \\
{\small $^\spadesuit$ Department of Linguistics, UT Austin \quad $^\ddagger$Division of Linguistics and Multilingual Studies, Nanyang Technological University} \\
{\small $^\heartsuit$Dept.~of Language Science and Technology/Division of AI and the Humanities, Hong Kong Polytechnic University} \vspace{.3em}
\\
{\small $^*$Equal contribution. Correspondence to \texttt{hai.hu@polyu.edu.hk}}
}

\begin{document}
\maketitle
\begin{abstract}

This study analyzes the predictive power of LM-derived, token-level surprisal on Mandarin Chinese reading times.
Using a suite of Chinese-Pythia models (14M--1.4B) trained from scratch with 30B tokens, we examine how well surprisal predicts first fixation duration, gaze duration, and total reading time in three paragraph-level eye-tracking corpora of Mandarin Chinese (GECO-CN, HKP, and MECO).
Contrary to previous null findings, our results show that surprisal is predictive of Chinese reading times.
However, whether predictive power scales with model size and the amount of training is corpus-specific: 
bigger models predict reading times better in GECO-CN, whereas inverse scaling emerges in HKP and, at the largest sizes, in MECO.
Subsequently, we find that checkpoints whose surprisal remains closer to $n$-gram statistics are better predictors of reading in HKP and MECO, providing one possible explanation for the inverse scaling behavior. 
All in all, the predictive power of surprisal on Chinese reading time measurements is corpus-specific, which cautions against drawing conclusions about language model scale from a single corpus.

\end{abstract}

\section{Introduction}


Humans read text incrementally, allocating different time and effort to each word they encounter.
The exact time and effort spent for each word has been proposed to be affected by the word's predictability.
This proposal has been formalized by surprisal theory~\citep{hale2001probabilistic,levy2008expectation}, which states that the more unexpected a word is, the higher the effort required to process it.
Such expectedness can be quantified by the probability (or surprisal) of the word in question given its preceding context.

\blue{LM-derived surprisal has been studied extensively in English \citep[among others]{wilcox2020predictive,shain2022robust,oh2023larger-worse,pimentel2023effect,kuribayashi-etal-2025-llm-internally-human-like}, as well as in Japanese \citep{kuribayashi-etal-2021-lower-perp-not-always-human-like-japanese,kuribayashi-etal-2022-context-limitations-japanese}, Dutch \citep{meister2021revisiting,boeve2025systematic-dutch}, and multilingual comparisons \citep{xu2023linearity,wilcox2023surprisal-11languages,de-varda-2023scaling-cogmodeling}. For Mandarin Chinese, however, two recent studies reported no reliable predictive contribution of LM surprisal \citep{xu2023linearity,sun2025attention-chinese-sentence-reading}.}\footnote{\blue{Earlier work used surprisal from Chinese $n$-gram models to predict reading times in children \citep{fan2020reading}.}}
\blue{These findings raise the question of whether the relationship holds across Mandarin reading datasets. Reliance on a single corpus can limit the generalizability of conclusions about reading behavior in a particular language.}

\blue{A second open question concerns scaling: how does surprisal's predictive power change with model size and the amount of training? English studies have examined both dimensions using intermediate checkpoints \citep{oh-schuler-2023-transformer,oh2023larger-worse,oh-etal-2024-frequency-explains-inverse}. Work in other languages has primarily compared model sizes \citep{wilcox2023surprisal-11languages,de-varda-2023scaling-cogmodeling,boeve2025systematic-dutch}, using fewer controlled monolingual model suites comparable to Pythia \citep{biderman2023pythia} or OPT \citep{zhang2022opt}. Evaluating checkpoints throughout training makes it possible to distinguish effects of model size from the effects of training data within a model.}

\blue{We address these questions through a controlled comparison across three Mandarin eye-tracking corpora. We train and release Chinese-Pythia models with five parameter sizes, from 14M to 1.4B, trained from scratch on approximately 30B tokens. Using 40 checkpoints per size, we examine first fixation duration (FFD), gaze duration (GD), and total reading time (TRT) with linear mixed-effects models. We also test one hypothesis about the scaling behavior of LMs on surprisal's fit to reading time, which is that alignment with local $n$-gram statistics is associated with its predictive power, following \citet{michaelov-levy-2026-ngram}. Specifically, we ask:}

\begin{enumerate}[itemsep=0pt, parsep=0pt, topsep=2pt]
    \item \blue{To what extent does LM surprisal predict Mandarin Chinese reading times across corpora and reading-time measures?}
    \item \blue{How does surprisal's predictive power vary with model size and the amount of training, and how do these patterns differ across corpora and measures?}
    \item \blue{How is the $n$-gram-likeness of LM surprisal related to its predictive power across training, and how does this relationship vary across corpora?}
\end{enumerate}

\blue{Our results show that surprisal improves reading-time fit on all three corpora, but the direction of scaling differs across them. Post-peak predictive power is associated with $n$-gram-likeness in HKP and MECO, while GECO-CN limits the generality of this account. Together, these results show why conclusions about LM scaling and reading-time prediction require comparisons across datasets.}

\section{Related Work}

\subsection{Surprisal and reading times in Chinese}

Most previous work on using information-theoretic metrics extracted from LMs to predict human reading time uses data collected in English~\citep{oh-clark-schuler-2022-comparison,oh2023larger-worse,pimentel2023effect,Shain2024}.
Although this line of inquiry has also expanded to other languages like Japanese~\citep{kuribayashi-etal-2021-lower-perp-not-always-human-like-japanese,kuribayashi-etal-2022-context-limitations-japanese} and Dutch~\citep{meister2021revisiting,boeve2025systematic-dutch}, Mandarin Chinese remains comparatively understudied.
The two existing studies found null results for LM surprisal's predictive power on reading times of Mandarin Chinese ~\citep{xu2023linearity,sun2025attention-chinese-sentence-reading}.
Specifically, \citet{xu2023linearity} studied data in six languages and found that surprisal does not add predictive power for eye-tracking measures for reading Mandarin Chinese.
\citet{sun2025attention-chinese-sentence-reading} reported that word surprisal is not a significant predictor, yielding only a weak effect on eye movements.
It is thus an open question whether Mandarin Chinese presents an outlier for the relationship between LM surprisal and eye-tracking measurements during reading.

\subsection{Inverse scaling of surprisal's predictive power}

While the predictive power of surprisal on English data has been widely replicated, an emerging line of work reports an inverse relationship between model size and the ability of surprisal to predict human reading times~\citep{oh2023larger-worse,oh-etal-2024-frequency-explains-inverse}.
Several explanations have been proposed for this ``bigger is not always better'' phenomenon~\citep{wilcox2025bigger-not-always-better}.
First, larger models appear to predict rare words too accurately:
they learn superhumanly complex associations for low-frequency lexical items, which diverge from human expectations and thereby reduce the correlation with reading times~\citep{oh-etal-2024-frequency-explains-inverse}.
\orange{Relatedly, \citet{michaelov-levy-2026-ngram} propose that reading times are shaped by simple $n$-gram statistics rather than the more complex statistics that large LMs learn.}
Second, the scaling mismatch may be driven by systematic biases tied to word class:
larger models tend to under-predict reading times for named entities while over-predicting them for function words, introducing a distortion that grows with model capacity~\citep{oh2023larger-worse}.
Third, this inverse scaling cannot be attributed to data leakage, as the effect persists even when models are trained on carefully filtered, leakage-free corpora~\citep{oh-etal-2025-not-due-to-data-leakage}.
Taken together, these findings suggest that as LMs scale up, their internal probability distributions become less human-like in reading-time prediction, and in turn, question the appropriateness of using even larger models as cognitive models of human processing.

Our study extends this debate by comparing corpora that differ in genre:
a detective novel (GECO-CN), news reports (HKP), and Wikipedia-style paragraphs (MECO).
If the relationship between surprisal and reading effort is genre-dependent, corpus-specific scaling trajectories should emerge---underscoring the importance of evaluating cognitive models across diverse linguistic materials.

\subsection{Systematic scaling beyond English}

A major obstacle to investigating scaling effects is the scarcity of model suites that are trained on identical data with publically available intermediate training checkpoints.
In English, models suites such as Pythia~\citep{biderman2023pythia} and OPT~\citep{zhang2022opt} has enabled controlled comparisons of surprisal estimates across model sizes and training steps.
However, comparable resources barely exist in other languages, and therefore researchers often rely on heterogeneous models with different architectures, training data, and tokenization schemes, which have confounded the scaling analyses.
In Dutch, for example, \citet{boeve2025systematic-dutch} compared multiple LMs of varying sizes and architectures, but the models did not share consistent training data, nor were the intermediate checkpoints released. Thus, a clean dissociation of model size from other factors could not be made.

Our work addresses this gap by training a suite of decoder-only models---which we refer to as Chinese-Pythia---with five model sizes (14M to 1.4B) on the same 30B-token Chinese corpus.
Because all models share the same architecture, tokenizer, and training data, we are able to systematically examine how surprisal's predictive power scales with model size and training data amount.
This allows us to test whether the inverse scaling behavior observed in English generalizes to a typologically distinct language and explore potential factors that moderate this relationship.

\section{Experimental Setup}
\subsection{Eye-tracking data}

\begin{table*}[t]
    \centering
    \small
    \setlength{\tabcolsep}{4.5pt}
    \begin{tabular}{l ccc cc cc cc}
        \toprule
        & & & & \multicolumn{2}{c}{\textbf{TRT}} & \multicolumn{2}{c}{\textbf{FFD}} & \multicolumn{2}{c}{\textbf{GD}} \\
        \cmidrule(lr){5-6} \cmidrule(lr){7-8} \cmidrule(lr){9-10}
        Corpus & $N_{\text{sbj}}$ & $N_{\text{words}}$ & PPL & $N_{\text{obs}}$ & Mean RT & $N_{\text{obs}}$ & Mean RT & $N_{\text{obs}}$ & Mean RT \\
        \midrule
        GECO-CN & 30 & 46,680 & 60.9 & 348,105 & 281.8 (196.3) & 348,527 & 206.5 (89.6)  & 253,298 & 221.4 (115.7) \\
        HKP     & 98 &  5,071 & 46.6 & 195,840 & 289.7 (202.1) & 196,048 & 209.8 (91.8)  & 158,274 & 222.4 (114.1) \\
        MECO    & 39 &  1,812 & 31.5 &  27,192 & 358.5 (256.1) &  26,293 & 227.0 (101.9) &  26,283 & 248.5 (135.5) \\
        \bottomrule
    \end{tabular}
    \caption{Descriptive statistics after preprocessing (Appendix~\ref{sec:app:eye-tracking:datasets}).
    $N_{\text{sbj}}$: number of subjects; $N_{\text{words}}$: number of aligned word positions (Section~\ref{sec:alignment}) with at least one TRT observation;
    PPL: perplexity under Qwen2.5-0.5B \citep{qwen25};
    $N_{\text{obs}}$: number of subject-by-word observations;
    Mean RT: mean reading time (SD) in ms.}
    \label{tab:corpus_combined}
\end{table*}

We analyze three large-scale eye-tracking datasets of naturalistic Mandarin Chinese reading---the Ghent Eye-tracking COrpus of sentence reading for Chinese-English bilinguals (GECO-CN; \citealp{GECO-CN}), the Hong Kong Corpus of Chinese Sentence and Passage Reading (HKP, we use the subset of data from paragraph reading rather than sentence reading; \citealp{Wu2023-HKC}), and the Multilingual Eye-Movement Corpus (MECO; \citealp{siegelman_wave_2025}).\footnote{
Details of the raw corpus are presented in Appendix \ref{sec:app:eye-tracking:datasets}.
}
GECO-CN contains eye movements from native Chinese speakers reading a detective novel translated into simplified Chinese.
HKC comprises reading data of simplified Chinese texts derived from Xinhua news reports.
MECO contains eye-tracking data collected using Wikipedia-style short passages in simplified and traditional Chinese.
To ensure comparability across corpora, we included only the simplified-Chinese, passage-reading portion of the data from all three sources.
Several commonly used Chinese reading time corpora were excluded because they either do not provide all eye-tracking metrics \citep{pan2022beijing-sentence-corpus}, or their stimuli are sentence-level materials used in controlled psycholinguistic experiments \citep{zhang2022database-cas-eye-tracking-corpus}.

We analyze three standard measurements of reading time \citep{rayner1998eye}:
\begin{enumerate*}[label=(\arabic*)]
    \item first fixation duration (FFD), the duration of the initial fixation on an area of interest during its first pass;
    \item gaze duration (GD), the sum of all first-pass fixations on the area of interest; and
    \item total reading time (TRT), the cumulative duration of all fixations on the area of interest across all passes.
\end{enumerate*}
Following \citet{smith2013effect} and \citet{xu2023linearity}, we preprocessed each reading-time measure by excluding skipped words (0\,ms reading time), extreme outliers (>2000\,ms), and paragraph-initial words.
We did not apply a hard surprisal cutoff because that would disproportionately remove multi-token Chinese words.\footnote{Observation counts remaining after each filtering step are listed in Table~\ref{tab:corpus:filter} of Appendix \ref{sec:app:eye-tracking:datasets}.}

Table~\ref{tab:corpus_combined} summarizes the three corpora after filtering.
Notably, MECO is read most slowly despite having the lowest perplexity, whereas GECO-CN has the highest perplexity and the fastest mean reading times.

\subsection{Models}
We train a series of decoder-only models in Mandarin Chinese, following the Pythia model suite~\citep{biderman2023pythia}, with similar configurations.
Our Chinese-Pythia suite includes models with the following parameter sizes:
14M, 70M, 160M, 410M and 1.4B.
We use the tokenizer from Chinese-Llama~\citep{cui2024-chinese-llama}.
The training data contains 30B tokens and is a subset of the CCI3.0-HQ corpus~\citep{wang2024cci3-hq}, which is a deduplicated collection of Chinese web text covering news, social media, and blogs.
We set the context length to 256 and the number of tokens trained per step to 40,960.
For each model size, we analyze 40 saved training checkpoints:
the 14 early checkpoints saved according to a doubling schedule (after $2^0$ to $2^{13}$, or 1 to 8{,}192 training steps), the checkpoint after 10{,}000 steps, and the 25 checkpoints saved after every 30{,}000 steps (from 30{,}000 to 750{,}000).
Together these span from the first training step to the end of the 30B-token run.
All model training was conducted on two A100 GPUs.\footnote{Our models will be publicly released upon publication.}

\subsection{Estimating surprisal from LMs}

We follow the literature in using the negative log probability of a token as its surprisal \cite{oh2023larger-worse,oh-schuler-2023-transformer,wilcox-etal-2023-language-model-quality-correlates-with-predictive-power,pimentel2023effect,wilcox2023surprisal-11languages}.
That is, for a token $x_t$ at index $t$ with preceding context $x_{<t}$, its surprisal is defined as:\footnote{We modify the scripts from \citet{haller-etal-2024-language} \url{https://github.com/DiLi-Lab/LM-cog-profiles} to compute the surprisal in Chinese.}
\[
S(x_t | x_{<t}) \equiv - \log p(x_t|x_{<t}).
\]
For a multi-token word $w=\{x_t,x_{t+1},...,x_n\}$, we sum up the negative log probilities of the tokens:
\[
S(w) = \sum_{i=t}^{n} - \log p(x_i|x_{<i}).
\]

\subsection{Aligning tokens between LMs and eye-tracking data}\label{sec:alignment}

\begin{table*}[t]
\centering
\small
\renewcommand{\arraystretch}{1.1}
\begin{CJK*}{UTF8}{gbsn} 
\begin{tabular}{@{}rl@{}}
\toprule
\multicolumn{2}{@{}l}{\textsc{Chinese}
  \enspace\textit{HKP eye-tracking corpus\,$+$\,Chinese-LLaMA tokenizer}} \\
\midrule
Corpus &
  国内 $\mid$
  \colorbox{cyan!20}{企业 $\mid$ 的} $\mid$
  \colorbox{red!15}{老总} $\mid$
  们 $\mid$ 也 $\mid$ 以 $\mid$ 同样 $\mid$ 热情 $\mid$ 关注 $\mid$
  \colorbox{orange!20}{这些 $\mid$ 人才} \\
LM &
  国内 $\mid$
  \colorbox{cyan!20}{企业的} $\mid$
  \colorbox{red!15}{老 $\mid$ 总} $\mid$
  们 $\mid$ 也 $\mid$ 以 $\mid$ 同样 $\mid$ 热情 $\mid$ 关注 $\mid$
  \colorbox{orange!20}{这些人 $\mid$ 才} \\
Aligned &
  国内 $\mid$
  \colorbox{cyan!20}{企业的} $\mid$
  \colorbox{red!15}{老总} $\mid$
  们 $\mid$ 也 $\mid$ 以 $\mid$ 同样 $\mid$ 热情 $\mid$ 关注 $\mid$
  \colorbox{orange!20}{这些人才} \\
  Gloss & Domestic $\mid$ company-DE $\mid$ leaders $\mid$ PL $\mid$ also $\mid$ with $\mid$ same $\mid$ enthusiasm $\mid$ focus-on $\mid$ these talents \\
  Trans. & Leaders from domestic companies also focus on these talents with the same enthusiasm.  \\
\bottomrule
\end{tabular}

\vspace{0.2em}
\footnotesize
\renewcommand{\arraystretch}{1.1}
\begin{tabular}{@{}lccc@{}}
\toprule
Mismatch type & Corpus tokens & LM tokens & Alignment operation \\
\midrule
\colorbox{red!15}{\strut LM finer\,} &
  老总 & 老 $\mid$ 总 & Sum LM surprisals \\
\colorbox{cyan!20}{\strut Corpus finer\,} &
  企业 $\mid$ 的 & 企业的 & Sum TRTs; Take first for FFD; Skip for GD \\
\colorbox{orange!20}{\strut Crossing\,} &
  这些 $\mid$ 人才 & 这些人 $\mid$ 才 & Merge both sides \\
\bottomrule
\end{tabular}
\caption{Alignment between eye-tracking corpus segmentation and LM tokenization.
Token boundaries are denoted using ``$\mid$''.
Colors mark the three types of boundary mismatch; each is resolved by aligning to the smallest span on which both segmentations agree, through the operations listed in the lower panel.}
\label{tab:alignment}
\end{CJK*}
\end{table*}

Aligning LM tokens with the word segmentation of an eye-tracking corpus is more complicated in Mandarin Chinese than in languages that have whitespaces, such as English and Dutch.
In those languages, an LM's subword tokens almost always nest within whitespace words, so the surprisal of a word is simply the sum of the surprisal of its tokens.
In Mandarin Chinese, however, the corpus segmentation and the LM tokenizer segment the same character string independently, so a word in the corpus and an LM token can coincide, the LM tokenization can be finer, the corpus segmentation can be finer, or their boundaries can cross (Table~\ref{tab:alignment}; see Table~\ref{tab:alignment_stats} for the frequency of each condition).
We therefore align the two sequences by identifying the shortest spans whose boundaries are shared by both segmentations, and compute surprisal and reading times for each span.

On the LM side, the surprisal of a span is the sum of the surprisal of its LM tokens.
Therefore, when the LM tokenization is finer, the span is a single corpus word, whose reading times are used as they are.
On the eye-tracking side, spans that contain more than one corpus word require combining word-level reading times, and the three measures differ in whether this is possible.
TRT, the total time spent on a region across all passes, is additive, so the TRT of a span is the sum of the TRTs of its words.
FFD, the duration of the first fixation on the span, equals the FFD of whichever of its words was fixated first, so we take the FFD of that word.
GECO-CN and HKP provide the information needed to identify this word, but MECO does not, so multi-word spans are excluded from the FFD analysis of MECO.
GD, the sum of first-pass fixations on the span, cannot be reconstructed from the GDs of its words, because it depends on whether the reader left the span between fixating them, which the corpora do not record.\footnote{For example, if a reader fixated B and then A in a span ABC, GD(ABC) is GD(B) + GD(A) if they fixated a word outside the span right after, but GD(B) + GD(A) + GD(C) if they fixated C right after.}
We therefore exclude the GDs of multi-word spans, which make up 21.7\% of aligned spans in GECO-CN, 14.1\% in HKP, and 10.0\% in MECO (See Table~\ref{tab:alignment_stats}).

\begin{table*}[t]
\centering
\resizebox{\textwidth}{!}{%
\begin{tabular}{lrrrrrrrrrr}
\toprule
Corpus & \#Subj & \#Units & \#Char & Unit Len & \#Align & No Merge & Corp. Finer & LM Finer & Crossing & C/L Ratio \\
\midrule
GECO-CN & 30 & 540 & 101,233 & 187.5 & 47,996 & 50.4\% & 15.0\% & 27.9\% & 6.7\% & 0.54 \\
HKP     & 98 &  41 &   9,066 & 221.1 &  5,117 & 79.7\% & 13.3\% &  6.2\% & 0.8\% & 2.16 \\
MECO    & 39 &  12 &   3,836 & 319.7 &  1,836 & 62.9\% &  6.6\% & 27.1\% & 3.4\% & 0.24 \\
\bottomrule
\end{tabular}}
\caption{Tokenization alignment breakdown. \#Units is the number of text units submitted to alignment (chunks of the GECO-CN novel, passage chunks of HKP, and paragraphs of MECO). \#Align is the total number of aligned word spans; the four condition counts sum to \#Align. C/L Ratio is the ratio of corpus-finer to LM-finer spans.}
\label{tab:alignment_stats}
\end{table*}

\subsection{Regression models}

We fit linear mixed-effects models to human reading data, with measures for reading time as the dependent variable, and word length, word frequency, and optionally surprisal as the independent variables.
All predictors are standardized, and the predictors of the current word are accompanied by their spillover terms from the preceding word.
The baseline model includes word length, log-transformed word frequency, and their spillover terms, with by-subject random slopes for word length; the surprisal model adds standardized surprisal and its spillover term.
We compute the difference in log-likelihood ($\Delta llh$) between the two models to measure the predictive power of surprisal.\footnote{See Appendix~\ref{sec:app:regression} for the full model specifications.}
Because $\Delta llh$ is uniformly positive across all 40 checkpoints for every model size, corpus, and measure (Section~\ref{sec:results}), we treat its magnitude, rather than per-fit significance, as the unit of comparison.

\subsection{The $n$-gram-likeness of LM surprisal}

This analysis follows \citet{michaelov-levy-2026-ngram}, who showed that Pythia checkpoints with stronger correlations to unigram, bigram, and trigram surprisal also tended to show stronger correlations with eye-tracking measures in the Provo Corpus \citep{luke2018provocorpuslarge}.
We ask whether the same qualitative pattern holds for Mandarin Chinese reading times.
If eye-movement measures are especially sensitive to local distributional statistics, then LLM-to-RT correlations should be higher where LLM surprisal aligns with $n$-gram surprisal.

We use the FFD, GD, and TRT eye-tracking data from all three corpora, indexing progressively more cumulative aspects of eye-movement behavior during reading.
Unigram, bigram, and trigram surprisal were computed with the released Chinese Wikipedia KenLM model \citep{kenlm} from the \texttt{edugp/kenlm} repository.
This model has its own tokenization scheme: it scores SentencePiece pieces \citep{kudo-richardson-2018-sentencepiece} of normalized text, which differ from both the corpus words and the LM tokens.
We did not align these pieces as in Section~\ref{sec:alignment}; instead, each text unit (Table~\ref{tab:alignment_stats}) was tokenized as a whole, each piece was assigned to a single aligned span, and the $n$-gram surprisal of a span is the sum over its pieces.
%
%
\orange{For each checkpoint, we computed two Pearson correlations: LM-to-RT, between LM surprisal and a reading-time measure, and LM-to-$n$-gram, between LM surprisal and $n$-gram surprisal, which we call $n$-gram-likeness.}

\section{Results}
\label{sec:results}

\subsection{Surprisal improves RT predictions}

\begin{figure*}[t]
    \centering
    \includegraphics[width=0.95\linewidth]{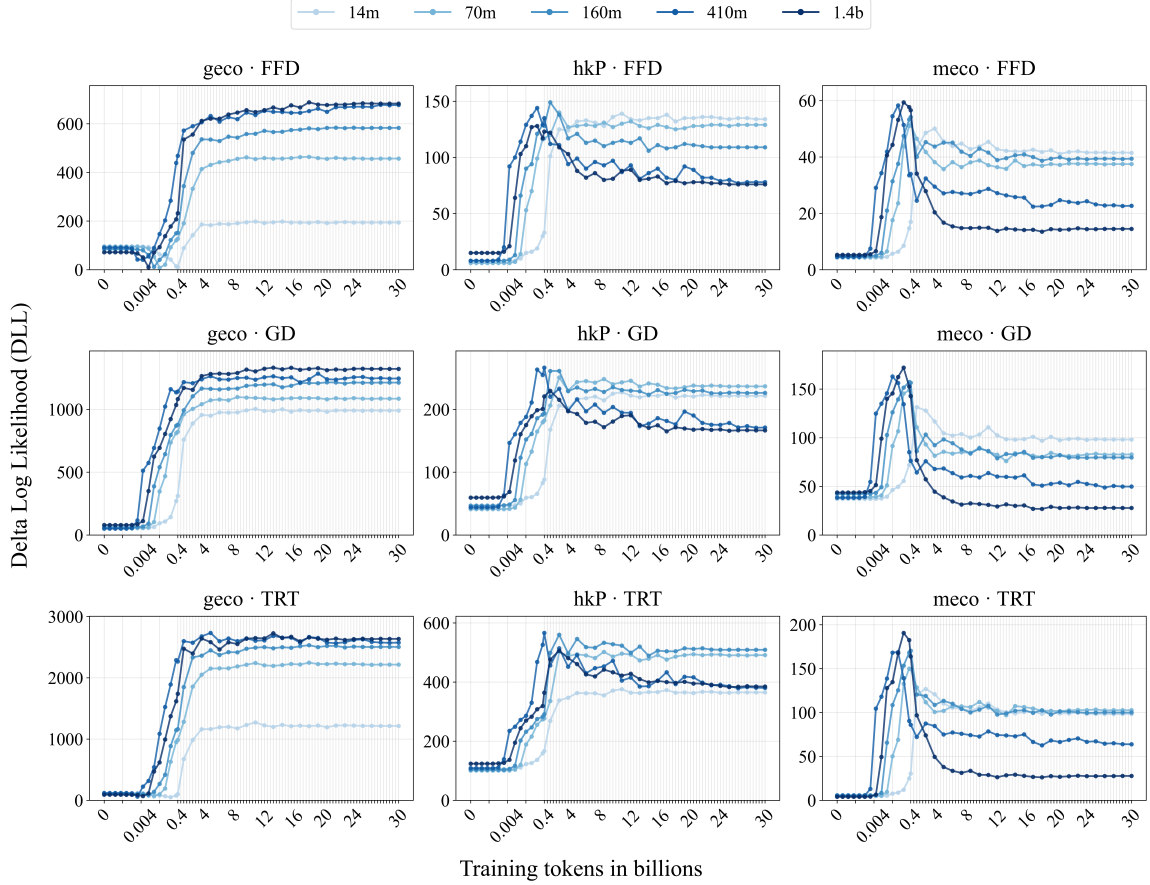}
    \caption{$\Delta llh$ for reading times across training. 
    Positive values indicate that including surprisal improves the prediction of reading times relative to the baseline.}
    \label{fig:dll}
\end{figure*}


\begin{table}[t]
\centering
\small
\setlength{\tabcolsep}{3pt}
\begin{tabular}{@{}llrrrrr@{}}
\toprule
Corpus & Measure & 14M & 70M & 160M & 410M & 1.4B \\
\midrule
\multirow{3}{*}{\textit{GECO}}
 & TRT & 1216 & 2219 & 2506 & 2610 & \textbf{2642} \\
 & FFD &  194 &  458 &  579 &  661 & \textbf{676} \\
 & GD  &  990 & 1086 & 1206 & 1249 & \textbf{1322} \\
\midrule
\multirow{3}{*}{\textit{HKP}}
 & TRT &  365 &  488 & \textbf{508} & 396 & 396 \\
 & FFD & \textbf{135} & 128 & 110 & 83 & 78 \\
 & GD  &  221 & \textbf{237} & 227 & 179 & 170 \\
\midrule
\multirow{3}{*}{\textit{MECO}}
 & TRT & 100 & \textbf{103} & 100 & 68 & 28 \\
 & FFD & \textbf{42} & 37 & 40 & 24 & 14 \\
 & GD  & \textbf{99} & 82 & 81 & 54 & 28 \\
\bottomrule
\end{tabular}
\caption{Mean $\Delta llh$ of the linear regression models over late-training checkpoints ($\geq$300k steps out of 750k steps in total), by corpus, measure, and model size. Bold marks the best-performing size per row. Higher values indicate that adding surprisal yields a better fit.}
\label{tab:dll-late}
\end{table}

Figure~\ref{fig:dll} presents the $\Delta\textit{llh}$ for each checkpoint, with corpora and reading time measures in different facets. Table~\ref{tab:dll-late} summarizes $\Delta\textit{llh}$ for late-training checkpoints.
The first clear pattern we observe is that including surprisal in the regression model improves predictive power relative to the baseline model, as evidenced by the positive $\Delta llh$ values throughout training, for all model sizes and on all three corpora.
Unlike findings from a previous study that found no such effect on a small Chinese eye-tracking dataset \citep{xu2023linearity}, our result aligns with the general pattern found on cross-linguistic studies, in particular the English ones \citep{oh-schuler-2023-transformer,oh2023larger-worse, wilcox2020predictive,wilcox2023surprisal-11languages,Shain2024}.

\subsection{Scaling with model size and training is corpus-specific}

We also observed strong corpus-specific effects in how predictive power scales with model size and training, as shown in Figure~\ref{fig:dll}.
In GECO-CN, $\Delta llh$ values increase sharply as model size and training data increase, and then reach a high plateau within about 1-4 billion training tokens;
the largest model provides the best predictor for every measure (TRT $\Delta llh$: 1216 for 14M vs.\ 2642 for 1.4B), and the top-end gain from 410M to 1.4B is small.
Better fit to human reading times is thus obtained by the more powerful models---those with more parameters and trained on more data.

However, HKP shows the opposite pattern, exhibiting inverse scaling with respect to both training tokens and model size:
for every model above 14M, predictive power peaks after about 0.2--2.5 billion training tokens and declines with further training, and late-training TRT $\Delta llh$ peaks at 508 for 160M and falls to 396 for both 410M and 1.4B;
for FFD and GD the two largest models provide the weakest predictors of all.

In MECO, the divergence between smaller and larger models is even more evident.
\orange{Larger models reach the highest peaks early in training (Figure~\ref{fig:dll}), but from about 1.2B training tokens (step 30k) onward, the 1.4B model fits worse than the 14M model at every checkpoint (late-training TRT $\Delta llh$: 100 at 14M vs.\ 28 at 1.4B; GD: 99 vs.\ 28), inverting the size ordering at the largest sizes---a clear bigger-is-worse effect once models are sufficiently trained.}

\subsection{Early versus late reading-time measures}

Across eye-tracking measures, FFD shows the weakest gains from surprisal, as reflected in its smaller $\Delta llh$ values (see Figure~\ref{fig:dll} and Table~\ref{tab:dll-late}).
This suggests that surprisal is less predictive of early fixation behavior than of later reading-time measures.

Besides, on HKP, scaling up LMs decreases fit to early measures more than late measures:
FFD $\Delta llh$ falls by 42\% from 14M to 1.4B (from 135 to 78), whereas TRT $\Delta llh$ is slightly higher at 1.4B than at 14M (396 vs.\ 365).
On MECO, by contrast, larger models lose predictive power on all three measures (FFD from 42 to 14, GD from 99 to 28, TRT from 100 to 28),
and on GECO-CN, FFD continues to improve with model size alongside later measures.

\subsection{\blue{$n$-gram alignment and reading-time prediction}}
\label{sec:ngram}

\begin{figure*}[t]
\centering
\includegraphics[width=\textwidth]{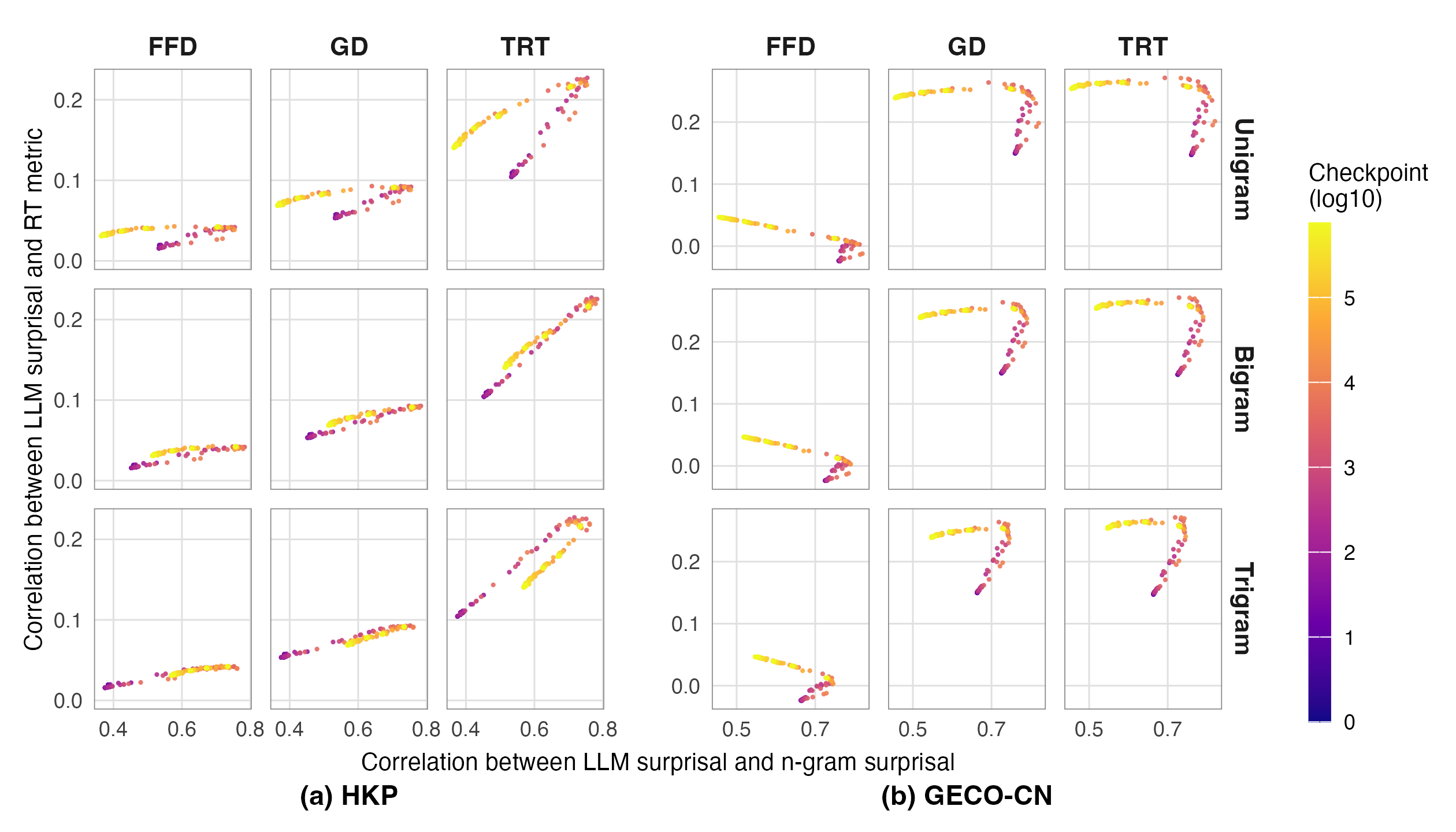}
\caption{\blue{Correspondence between $n$-gram-likeness and LM--RT correlation in (a) HKP and (b) GECO-CN. Each point represents one checkpoint from one of the five model sizes. The horizontal axis is the Pearson correlation between LM and $n$-gram surprisal; the vertical axis is the Pearson correlation between LM surprisal and a reading-time measure. Rows distinguish $n$-gram orders, columns distinguish reading-time measures, and color indicates the training step on a $\log_{10}$ scale.}}
\label{fig:exp2-correspondence}
\end{figure*}

\begin{table*}[t]
\centering
\small
\setlength{\tabcolsep}{7pt}
\setlength{\tabcolsep}{5pt}
\begin{tabular}{@{}lrrrrrrr@{}}
\toprule
\multicolumn{8}{@{}l}{\textit{(a) Training changes for the 1.4B model}} \\
Corpus & Drop in trigram-likeness & \multicolumn{3}{c}{Step of peak $\Delta llh$} & \multicolumn{3}{c}{Peak-to-final $\Delta llh$ loss} \\
\cmidrule(lr){3-5} \cmidrule(lr){6-8}
& & FFD & GD & TRT & FFD & GD & TRT \\
\midrule
GECO-CN & 0.19 & 450,000 & 330,000 & 330,000 & 1\% & 1\% & 3\% \\
HKP & 0.19 & 4,096 & 30,000 & 60,000 & 41\% & 27\% & 24\% \\
MECO & 0.32 & 4,096 & 4,096 & 4,096 & 76\% & 84\% & 85\% \\
\bottomrule
\end{tabular}
\setlength{\tabcolsep}{7pt}

\medskip
\begin{tabular}{@{}lrrrrrrrrr@{}}
\toprule
\multicolumn{10}{@{}l}{\textit{(b) $n$-gram surprisal and word-level mean reading time: Pearson $r$}} \\
& \multicolumn{3}{c}{GECO-CN} & \multicolumn{3}{c}{HKP} & \multicolumn{3}{c}{MECO} \\
\cmidrule(lr){2-4} \cmidrule(lr){5-7} \cmidrule(lr){8-10}
$n$-gram & FFD & GD & TRT & FFD & GD & TRT & FFD & GD & TRT \\
\midrule
Unigram & $-$0.04 & 0.13 & 0.41 & 0.14 & 0.27 & 0.44 & 0.03 & 0.25 & 0.38 \\
Bigram  & $-$0.01 & 0.15 & 0.43 & 0.20 & 0.32 & 0.47 & 0.09 & 0.29 & 0.41 \\
Trigram & 0.00 & 0.15 & 0.42 & 0.21 & 0.33 & 0.47 & 0.11 & 0.29 & 0.39 \\
\bottomrule
\end{tabular}
\caption{\blue{Two cross-corpus comparisons for the $n$-gram analysis. (a) An illustration at 1.4B parameters: trigram-likeness is the Pearson correlation between LM and trigram surprisal. Each decline is measured from that quantity's own maximum to the final checkpoint (step 750,000), so the two declines need not cover the same interval. Trigram-likeness drops are absolute differences in $r$ and are shared across the three reading-time measures. Peak steps and losses are computed separately for FFD, GD, and TRT from the saved regression summaries, using the earliest checkpoint in a tie; losses are percentages of peak $\Delta llh$, rounded to whole percentages. (b) Direct correlations of $n$-gram surprisal with reading times averaged over subjects at each word, rather than correlations across checkpoints.}}
\label{tab:ngram-summary}
\end{table*}


\blue{Following \citet{michaelov-levy-2026-ngram}, we examine the relations between three quantities: the raw LM--RT correlation at each checkpoint, $n$-gram-likeness (the correlation between LM and $n$-gram surprisal), and regression-based predictive power ($\Delta llh$).
We first ask whether checkpoints with greater $n$-gram-likeness also have stronger LM--RT correlations, then examine their relationship with $\Delta llh$. See details in Appendix~\ref{sec:app:ngram-corr}.}

\paragraph{\blue{$n$-gram-likeness and LM--RT correlation.}}
\blue{Figure~\ref{fig:exp2-correspondence} plots $n$-gram-likeness and LM--RT correlations across checkpoints in HKP and GECO-CN.
We again see corpus-specific effects, with markedly different patterns in HKP and GECO-CN.
In HKP, across the 200 checkpoints pooled over model sizes, LM--RT correlations track bigram- and trigram-likeness closely ($r = 0.85$--$0.99$); that is, they rise and fall in tandem (Figure~\ref{fig:exp2-correspondence}a).
MECO, which is not plotted, shows a similar pattern for GD and TRT, but not FFD.
However, in GECO-CN, the pooled correlations are negative for all orders and measures ($r = -0.90$ to $-0.28$), as shown in Figure~\ref{fig:exp2-correspondence}b.}

\begin{table*}[t]
\centering
\small
\setlength{\tabcolsep}{4pt}
\begin{tabular}{@{}llrrrrrrrrrr@{}}
\toprule
& & \multicolumn{2}{c}{14M} & \multicolumn{2}{c}{70M} & \multicolumn{2}{c}{160M} & \multicolumn{2}{c}{410M} & \multicolumn{2}{c}{1.4B} \\
\cmidrule(lr){3-4} \cmidrule(lr){5-6} \cmidrule(lr){7-8} \cmidrule(lr){9-10} \cmidrule(lr){11-12}
Corpus & Measure & all & post & all & post & all & post & all & post & all & post \\
\midrule
GECO-CN & FFD & 0.65 & $-$0.42 & $-$0.70 & $-$0.24 & $-$0.88 & $-$0.46 & $-$0.92 & 1.00 & $-$0.92 & $-$0.15 \\
 & GD & 0.93 & $-$0.69 & $-$0.32 & 0.48 & $-$0.64 & 0.05 & $-$0.67 & 0.33 & $-$0.70 & 0.33 \\
 & TRT & 0.86 & $-$0.76 & $-$0.55 & $-$0.37 & $-$0.79 & $-$0.60 & $-$0.81 & 0.51 & $-$0.81 & 0.49 \\
\midrule
HKP & FFD & 0.98 & 0.50 & 0.98 & 0.70 & 0.98 & 0.93 & 0.98 & 0.95 & 0.98 & 0.90 \\
 & GD & 0.99 & 0.48 & 0.96 & 0.48 & 0.92 & 0.85 & 0.91 & 0.93 & 0.92 & 0.94 \\
 & TRT & 0.98 & 0.66 & 0.89 & 0.35 & 0.79 & 0.66 & 0.73 & 0.87 & 0.66 & 0.95 \\
\midrule
MECO & FFD & 0.98 & 0.92 & 0.27 & 0.94 & $-$0.06 & 0.78 & 0.35 & 0.98 & 0.66 & 1.00 \\
 & GD & 0.99 & 0.95 & 0.66 & 0.91 & 0.54 & 0.93 & 0.74 & 0.98 & 0.90 & 0.99 \\
 & TRT & 0.98 & 0.95 & 0.30 & 0.92 & 0.15 & 0.96 & 0.36 & 0.97 & 0.70 & 1.00 \\
\bottomrule
\end{tabular}
\caption{\blue{Pearson correlations between trigram-likeness and $\Delta llh$ across checkpoints within each model size. \textit{All} uses all 40 checkpoints; \textit{post} uses checkpoints from the measure-specific $\Delta llh$ peak through the final checkpoint, including the peak. Positive values indicate that more trigram-like checkpoints tend to have higher predictive power. Checkpoints are dependent; the GECO-CN FFD post-peak value at 410M uses only three checkpoints.  Table~\ref{tab:ngram-dll-r} in the Appendix provides all three $n$-gram orders and checkpoint counts.}}
\label{tab:trigram-dll-r}
\end{table*}

\paragraph{\blue{$n$-gram-likeness and $\Delta llh$.}}
\blue{Across the full training trajectory, greater trigram-likeness is associated with higher predictive power in HKP and, with one exception, MECO, but with lower predictive power in GECO-CN above 14M. The \textit{all} columns in Table~\ref{tab:trigram-dll-r} report correlations between trigram-likeness and $\Delta llh$ across all 40 checkpoints, separately for each corpus, model size, and reading-time measure. All correlations are positive in HKP. In MECO, they are positive except for the near-zero correlation for 160M FFD ($r = -0.06$). In GECO-CN, they are negative for every measure above 14M, but positive at 14M. Thus, the overall relationship between trigram-likeness and predictive power differs across corpora.}

\blue{We then examine post-peak checkpoints specifically to ask whether declining predictive power accompanies declining trigram-likeness. The \textit{post} columns in Table~\ref{tab:trigram-dll-r} use checkpoints from each model's measure-specific $\Delta llh$ peak through the end of training, including the peak. All correlations are positive in HKP and MECO, consistent with the two quantities declining together during later training. GECO-CN instead shows mixed signs: eight correlations are negative and seven are positive. Its negative full-training relationship therefore does not consistently extend to the post-peak interval.} 

\blue{Together, the two analyses show a consistently positive association in HKP and a predominantly positive association in MECO. GECO-CN differs: the association is negative across the full trajectory above 14M, but inconsistent after the peak.}

\paragraph{\blue{Cross-corpus differences in the role of local statistics.}}
\blue{The relationship between change in trigram-likeness and decrease in predictive power also differs across corpora.
As Table~\ref{tab:ngram-summary}a shows for the 1.4B model, GECO-CN and HKP exhibit the same peak-to-final drop in trigram-likeness, but GECO-CN's $\Delta llh$ values peak much later (at step 330,000 or later) and decline much less.
MECO exhibits the largest decline in both quantities.
Declining trigram-likeness alone therefore cannot explain the cross-corpus differences in decrease in predictive power.
Table~\ref{tab:ngram-summary}b provides a complementary comparison: direct associations between $n$-gram surprisal and word-level mean reading times are weaker for GECO-CN's early measures, particularly FFD, but TRT associations are similar across corpora ($r = 0.38$--$0.47$).
Weak early-measure associations thus cannot account for the full contrast in predictive power.
Mean trigram surprisal under the Wikipedia model is also higher in GECO-CN (5.3 nats per character) than in HKP (4.3) or MECO (4.2), suggesting a possible resource mismatch without establishing its role in these differences.}


\section{General Discussion}

Adding LM-derived surprisal to baseline mixed-effects regression models improved fit for every reading time measure on all three corpora.
LM surprisal is thus predictive of Mandarin Chinese reading times, contrary to earlier null findings.
How this predictive power scales with model size and training, however, differed across corpora: it grew with scale in GECO-CN and inverted in HKP and MECO.
Below, we relate our results to earlier studies of Mandarin Chinese (Section~\ref{sec:discussion:prior}), characterize the corpus-specific scaling (Section~\ref{sec:discussion:scaling}), \blue{assess the scope and limits of the $n$-gram account (Section~\ref{sec:discussion:mechanisms})}, and assess text genre and reading regime as potential moderators (Section~\ref{sec:discussion:genre}).

\subsection{Relation to earlier null findings}
\label{sec:discussion:prior}

Earlier studies found no reliable surprisal effect on Chinese reading times \citep{xu2023linearity,sun2025attention-chinese-sentence-reading}.
This discrepancy likely stems from model and sample differences: \citet{xu2023linearity} analyzed small datasets using GPT-2 and mGPT (neither trained primarily on Chinese), whereas we use Chinese-trained LMs and three corpora totaling >570,000 TRT observations.

The results are also less contradictory than they appear.
In HKP, raw correlations between surprisal and reading times are low (at most 0.04 for FFD, 0.09 for GD, and 0.23 for TRT), far below those in English \citep{michaelov-levy-2026-ngram}.
Because the effect is weak, analyses using small samples or non-Mandarin LMs may not able to detect it.
Finally, because Mandarin uses an unspaced writing system with ambiguous word boundaries, corpus-defined words may align less well with the units of predictability that drive eye movements, compared to English or Dutch.

\subsection{The direction of scaling is corpus-specific}
\label{sec:discussion:scaling}

The three corpora show a similar trend in early LM training: fit to reading times rises steeply and peaks or plateaus within the first few billion tokens, with larger models generally reaching higher values (see Figure~\ref{fig:dll}), although on HKP the 1.4B model peaks below the 160M and 410M models.
They show divergent trends with further training.
On GECO-CN, the fit plateaus and the size ordering holds, with diminishing gains at the top; on HKP, fit declines after the peak, most steeply for the largest models, such late-training fit is best at small or intermediate sizes; and on MECO, the decline is the steepest and inverts the size ordering at the largest sizes (Section~\ref{sec:results}).
As the $\Delta llh$ curves have converged by the end of training, these patterns are asymptotic rather than artifacts of under-training.
On HKP, the 1.4B model's fit to TRT peaks at step 60k (about 2.5B tokens), close to the roughly two billion tokens at which Transformer surprisal best predicts English reading times \citep{oh-schuler-2023-transformer}; checkpoints at this stage fit better than fully trained ones on HKP (above 14M) and MECO, but not on GECO-CN.

The eye-tracking measures show different trends as well.
In HKP, scale mainly worsens fit to early measures: from 14M to 1.4B, FFD $\Delta llh$ drops by 42\%, whereas TRT $\Delta llh$ rises slightly (see Table~\ref{tab:dll-late}), echoing \citet{de-varda-2023scaling-cogmodeling}, who found across several languages that larger models predict early eye-tracking measures worse but late measures better.
In MECO, however, all three measures drop by two-thirds or more, so this asymmetry is not general.

These results bear on the scaling debate in English, where LM quality improves predictive power for smaller models and training sets \citep{wilcox2020predictive, wilcox-etal-2023-language-model-quality-correlates-with-predictive-power} but inverse scaling holds for larger Transformers \citep{oh-clark-schuler-2022-comparison, oh2023larger-worse}, inviting an explanation in terms of model regime.
Our models show both patterns over the same 14M--1.4B range, depending on corpus and measure, so part of this disagreement may reflect the effects of the reading materials and eye-tracking measures rather than model regime alone.

\subsection{\blue{Local statistics and the limits of the $n$-gram account}}
\label{sec:discussion:mechanisms}

\blue{The post-peak associations in HKP and MECO are consistent with the proposal that sensitivity to local distributional statistics contributes to reading-time prediction \citep{michaelov-levy-2026-ngram}. One interpretation is that early training captures these statistics, whereas later training learns higher-order regularities that improve next-word prediction without improving fit to reading times. Related accounts emphasize frequency and the learning of rare words or entities \citep{ohschuler2025frequency,opedal2024role}. Our correlations support an association between $n$-gram alignment and predictive power, but also shows its limitation.}

\blue{Specifically, the results in Table~\ref{tab:ngram-summary}a show that a similar decline in trigram-likeness can accompany very different losses of predictive power for eye-tracking measures. Moving away from $n$-gram statistics is therefore not sufficient to explain the decline in fit, and $n$-gram-likeness alone does not explain why scaling differs across corpora. The mixed post-peak correlations in GECO-CN also caution against treating its all-checkpoint negative association as a uniform relationship throughout training.}

\blue{Two properties of GECO-CN may qualify the account. Its higher mean trigram surprisal suggests a possible mismatch with the Wikipedia-trained resource, which was used to train the $n$-gram models.
Its weaker direct $n$-gram associations with FFD and GD may also help explain the early-measure differences (Table~\ref{tab:ngram-summary}b). However, direct TRT correlations are similar across corpora, so weak early-measure associations cannot explain the full contrast. Section~\ref{sec:discussion:genre} considers text genre and reading regime.}

\subsection{Text genre and reading regime}
\label{sec:discussion:genre}

One possibility is that expository text encourages skimming, loosening the link between predictability and eye movements.
MECO, the encyclopedic corpus, argues against this: it was read most slowly (mean TRT 359\,ms vs.\ 282\,ms in GECO-CN; See Table~\ref{tab:corpus_combined}), had the smallest share of TRT observations removed as skipped or missing (about 45\%, vs.\ 51\% in GECO-CN and 60\% in HKP; see Table~\ref{tab:corpus:filter}), and follows each text with comprehension questions.
HKP's news reports give mixed evidence: the most skipping, but slower reading than GECO-CN (mean TRT 290 vs.\ 282\,ms).
More fundamentally, a looser link would weaken surprisal effects overall rather than change the direction of scaling, as we observe in our analysis: surprisal from small models predicts reading times in all three corpora, and what differs is to what extent larger, longer-trained models improve on it.
Genre remains a candidate, since GECO-CN, the only narrative corpus, is also the only one whose late-training fit increases with model size, but with a single narrative corpus, genre is confounded with every other characteristic unique to GECO-CN.
\blue{Our data thus do not yet explain why scaling differs across corpora; the $n$-gram account is consistent with the post-peak associations in HKP and MECO but is limited by the GECO-CN pattern (Section~\ref{sec:discussion:mechanisms}).}

\section{Conclusion}

We trained and evaluated a suite of Chinese LMs to trace how the predictive power of surprisal on reading times changes with model size and amount of training data.
Surprisal improved fit in all cases, but its scaling behavior was corpus-specific: larger models predicted reading times better in GECO, whereas in HKP and MECO fit peaked early in training and then declined, most steeply for the largest models.
On all three corpora, surprisal drifted away from $n$-gram statistics after early training; this drift accompanied the decline in HKP and MECO but left fit intact on GECO.
Whether bigger or better-trained models predict reading times better thus depends on the materials, cautioning against drawing conclusions about scaling behavior from a single corpus.

\section{Limitations}

This study has three main limitations.
\orange{
First, the publicly available $n$-gram models that we used were trained on Wikipedia rather than on our training data.}
Second, $\Delta llh$ magnitudes scale with the number of observations and are not directly comparable across corpora;
all comparisons in this work are therefore made within a corpus, and simulation-based or normalized effect-size measures would be needed for genuine cross-corpus quantification.
Third, genre may moderate how surprisal's predictive power scales (Section~\ref{sec:discussion:genre}), but the present design cannot isolate its effect: GECO-CN is the only narrative corpus in this study, so genre is confounded with everything else specific to that corpus.

\section{Ethical Considerations}
This study uses eye-tracking data from previously published corpora and involves no new human-subject data collection. The Chinese-Pythia models were trained on publicly available text (CCI3-HQ) and will be released for research purposes.


\clearpage

\bibliography{tacl2021}
\bibliographystyle{acl_natbib}

\clearpage
\appendix

\section{Details on the three eye-tracking datasets}
\label{sec:app:eye-tracking:datasets}

\begin{itemize}
    \item Ghent Eye-tracking COrpus for Chinese-English bilinguals, \textbf{GECO-CN}~\citep{GECO-CN} contains eye-movement data from 30 native Chinese (L1) speakers reading the detective novel \textit{The Mysterious Affair at Styles} by Agatha Christie \citep{christie20}. Translation was revised by the first author, who is a native Chinese speaker. The published Chinese version of the novel comprises 59,403 words and 5,053 unique types; our extraction of GECO-CN comprises 895,938 observations.
    \item Hong Kong Corpus of Chinese sentence and passage reading, \textbf{HKC}~\citep{Wu2023-HKC} releases eye-tracking data of 98 L1 speakers of Chinese, who read 300 single-line sentences and 7 multiline passages selected from the XIN subcorpus of the Chinese Gigaword Corpus \citep{ma-chen-2003-introduction,ma-huang-2006-uniform}, with no overlap in between. The single-line sentences contain 5,150 words, while the passages include 4,967 words. Taken together, HKC comprises 10,117 words covering a variety of topics including celebrity news, city development, education, etc. In our extraction, the seven passages comprise 41 sentence lines and 5,911 words and 1,582 types.
    \item Multilingual Eye-Movement Corpus, \textbf{MECO}~\citep{siegelman_wave_2025} comprises eye-tracking data from native Chinese speakers reading 12 Wikipedia-like paragraphs; the published Chinese sample includes 77 participants reading texts in both simplified and traditional Chinese. We analyze the simplified-Chinese texts, which contains 71 sentences--2,024 words and 1,051 types.

\end{itemize}
Preprocessing proceeds in three steps, applied to each reading-time measure: (1) we exclude words with a reading time of 0 or missing; (2) we exclude words with non-finite surprisal or reading times of 2000 ms or longer; (3) we exclude paragraph-initial words. We did not apply surprisal cutoff as it would remove multi-token words disproportionally; words attached to punctuation and sentence-final words are retained, too.
See Table~\ref{tab:corpus:filter} for details.

\begin{table}[t]
    \centering
    \setlength{\tabcolsep}{4.5pt}
    \begin{tabular}{ll rrr}
        \toprule
        & & GECO-CN & HKP & MECO \\
        \midrule
        \multicolumn{2}{l}{Merged} & 719,518 & 489,847 & 49,730 \\
        \midrule
        TRT & Step 1 & 353,763 & 197,317 & 27,521 \\
            & Step 2 & 353,328 & 197,141 & 27,420 \\
            & Step 3 & 348,105 & 195,840 & 27,192 \\
        \addlinespace
        FFD & Step 1 & 353,763 & 197,355 & 26,527 \\
            & Step 2 & 353,761 & 197,352 & 26,527 \\
            & Step 3 & 348,527 & 196,048 & 26,293 \\
        \addlinespace
        GD  & Step 1 & 257,369 & 159,373 & 26,517 \\
            & Step 2 & 257,347 & 159,360 & 26,517 \\
            & Step 3 & 253,298 & 158,274 & 26,283 \\
        \bottomrule
    \end{tabular}
    \caption{Observations left after each preprocessing step, from the merged data after tokenization alignment (Section~\ref{sec:alignment}). Steps remove (1) skipped words, (2) non-finite surprisal and reading times of 2000\,ms or more, and (3) paragraph-initial words. Counts are identical for all models and checkpoints.}
    \label{tab:corpus:filter}
\end{table}

\begin{figure*}[t]
\centering
\includegraphics[width=\textwidth]{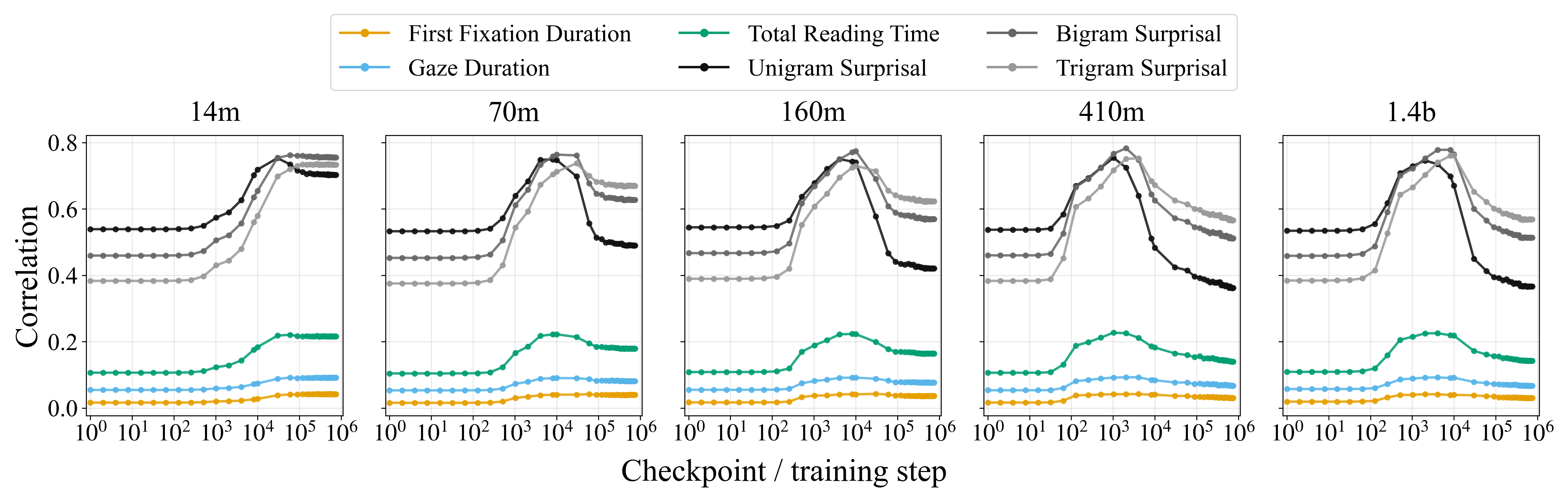}
\caption{\blue{Training trajectories of raw LM--RT correlations (colored lines) and $n$-gram-likeness (gray and black lines) in HKP. Each panel shows one Chinese Pythia model size; the horizontal axis is training step on a logarithmic scale and the vertical axis is Pearson $r$. These raw correlations are distinct from regression-based $\Delta llh$.}}
\label{fig:hkp-exp2-trajectories}
\end{figure*}

\section{Regression models used in this study}
\label{sec:app:regression}

Details of the regression models reported in the main section are presented below.\footnote{We modify the code base of \citet{xu2023linearity} to run the regression analyses in this study.  }  \texttt{RT} is the reading time (i.e., FFD, GD, TRT) of the word of interest (\( w_{i} \)). \texttt{surp.s} refers to the standardized surprisal of \( w_{i} \), while \texttt{wlen.s} and \texttt{logfreq.s} respectively denote the standardized word length and standardized log-transformed word frequency of \( w_{i} \); a suffix \texttt{1} (e.g., \texttt{surp1.s}, \texttt{wlen1.s}) denotes the corresponding spillover term for the preceding word. $\Delta llh$ refers to the change in log-likelihood, calculated as the difference between regression models with and without surprisal.
\paragraph{Predictive power of surprisal (Section \ref{sec:results})} For the tests assessing the effect of surprisal within each corpus, we use the following model:
\begin{multline*}
\textcolor{blue}{\texttt{lmer}}(\texttt{RT} \sim \texttt{surp.s} + \texttt{surp1.s} + \\
\texttt{wlen.s} + \texttt{wlen1.s} + \texttt{logfreq.s} + \\
\texttt{logfreq1.s} + (\texttt{wlen.s} | \texttt{subject}))
\end{multline*}

The baseline model is identical except that the surprisal terms are removed:
\begin{multline*}
\textcolor{blue}{\texttt{lmer}}(\texttt{RT} \sim \texttt{wlen.s} + \texttt{wlen1.s} + \\
\texttt{logfreq.s} + \texttt{logfreq1.s} + \\
(\texttt{wlen.s} | \texttt{subject}))
\end{multline*}
Both models are fit with maximum likelihood.

\section{Correlations in the $n$-gram analysis}
\label{sec:app:ngram-corr}

\blue{The raw LM--RT and LM--$n$-gram correlations in Section~\ref{sec:ngram} are computed separately for each checkpoint over merged subject--word observations, before the preprocessing used for regression (Table~\ref{tab:corpus:filter}). Skipped words count as 0\,ms for TRT and are excluded for FFD and GD. Figure~\ref{fig:hkp-exp2-trajectories} plots these training trajectories for HKP; Figure~\ref{fig:exp2-correspondence} shows their correspondence in HKP and GECO-CN. MECO's pooled associations are reported in Section~\ref{sec:ngram}.}

\blue{The subsequent correlations use checkpoints as observations. The pooled raw-correlation summaries in Section~\ref{sec:ngram} combine 200 checkpoints across five model sizes within each corpus. Table~\ref{tab:ngram-dll-r} instead correlates $n$-gram-likeness with $\Delta llh$ separately within each size, using either all 40 checkpoints or those from that size's measure-specific $\Delta llh$ peak onward. Tied peaks are resolved to the earlier checkpoint. Checkpoints are dependent, so these are descriptive summaries and no $p$-values are reported. Table~\ref{tab:ngram-summary}b is a distinct analysis: it correlates $n$-gram surprisal directly with word-level reading times averaged over subjects.}

\begin{table*}[t]
\centering
\setlength{\tabcolsep}{3.5pt}
\begin{tabular}{l l rr rr rr rr rr}
\toprule
& & \multicolumn{2}{c}{14M} & \multicolumn{2}{c}{70M} & \multicolumn{2}{c}{160M} & \multicolumn{2}{c}{410M} & \multicolumn{2}{c}{1.4B} \\
\cmidrule(lr){3-4} \cmidrule(lr){5-6} \cmidrule(lr){7-8} \cmidrule(lr){9-10} \cmidrule(lr){11-12}
Measure & $n$-gram & all & post & all & post & all & post & all & post & all & post \\
\midrule
\multicolumn{12}{l}{\textit{GECO-CN}} \\
FFD & Unigram & $-$0.84 & $-$0.07 & $-$0.99 & $-$0.42 & $-$0.99 & $-$0.58 & $-$0.99 & 0.79 & $-$0.99 & $-$0.01 \\
    & Bigram  & 0.21 & $-$0.35 & $-$0.95 & $-$0.28 & $-$0.97 & $-$0.59 & $-$0.98 & 0.95 & $-$0.98 & $-$0.01 \\
    & Trigram & 0.65 & $-$0.42 & $-$0.70 & $-$0.24 & $-$0.88 & $-$0.46 & $-$0.92 & 1.00 & $-$0.92 & $-$0.15 \\
    & $N$     & 40 & 17 & 40 & 11 & 40 & 8 & 40 & 3 & 40 & 11 \\
\addlinespace
GD  & Unigram & $-$0.52 & $-$0.35 & $-$0.84 & 0.46 & $-$0.87 & $-$0.16 & $-$0.86 & 0.38 & $-$0.87 & 0.42 \\
    & Bigram  & 0.63 & $-$0.61 & $-$0.73 & 0.47 & $-$0.81 & $-$0.16 & $-$0.79 & 0.39 & $-$0.82 & 0.39 \\
    & Trigram & 0.93 & $-$0.69 & $-$0.32 & 0.48 & $-$0.64 & 0.05 & $-$0.67 & 0.33 & $-$0.70 & 0.33 \\
    & $N$     & 40 & 17 & 40 & 19 & 40 & 9 & 40 & 10 & 40 & 15 \\
\addlinespace
TRT & Unigram & $-$0.64 & $-$0.62 & $-$0.95 & $-$0.53 & $-$0.96 & $-$0.58 & $-$0.94 & 0.55 & $-$0.94 & 0.60 \\
    & Bigram  & 0.51 & $-$0.74 & $-$0.88 & $-$0.39 & $-$0.92 & $-$0.61 & $-$0.90 & 0.54 & $-$0.90 & 0.56 \\
    & Trigram & 0.86 & $-$0.76 & $-$0.55 & $-$0.37 & $-$0.79 & $-$0.60 & $-$0.81 & 0.51 & $-$0.81 & 0.49 \\
    & $N$     & 40 & 17 & 40 & 11 & 40 & 11 & 40 & 22 & 40 & 15 \\
\midrule
\multicolumn{12}{l}{\textit{HKP}} \\
FFD & Unigram & 0.89 & 0.49 & $-$0.16 & 0.75 & $-$0.24 & 0.91 & 0.03 & 0.91 & 0.09 & 0.84 \\
    & Bigram  & 0.97 & 0.55 & 0.85 & 0.73 & 0.74 & 0.92 & 0.74 & 0.93 & 0.77 & 0.87 \\
    & Trigram & 0.98 & 0.50 & 0.98 & 0.70 & 0.98 & 0.93 & 0.98 & 0.95 & 0.98 & 0.90 \\
    & $N$     & 40 & 17 & 40 & 24 & 40 & 25 & 40 & 28 & 40 & 28 \\
\addlinespace
GD  & Unigram & 0.89 & 0.62 & $-$0.27 & 0.53 & $-$0.46 & 0.88 & $-$0.24 & 0.92 & $-$0.20 & 0.93 \\
    & Bigram  & 0.97 & 0.61 & 0.79 & 0.50 & 0.56 & 0.86 & 0.52 & 0.92 & 0.56 & 0.94 \\
    & Trigram & 0.99 & 0.48 & 0.96 & 0.48 & 0.92 & 0.85 & 0.91 & 0.93 & 0.92 & 0.94 \\
    & $N$     & 40 & 17 & 40 & 24 & 40 & 24 & 40 & 26 & 40 & 25 \\
\addlinespace
TRT & Unigram & 0.89 & 0.62 & $-$0.43 & 0.34 & $-$0.67 & 0.67 & $-$0.52 & 0.86 & $-$0.60 & 0.94 \\
    & Bigram  & 0.97 & 0.75 & 0.67 & 0.35 & 0.32 & 0.68 & 0.23 & 0.86 & 0.15 & 0.95 \\
    & Trigram & 0.98 & 0.66 & 0.89 & 0.35 & 0.79 & 0.66 & 0.73 & 0.87 & 0.66 & 0.95 \\
    & $N$     & 40 & 17 & 40 & 24 & 40 & 24 & 40 & 26 & 40 & 24 \\
\midrule
\multicolumn{12}{l}{\textit{MECO}} \\
FFD & Unigram & $-$0.64 & 0.91 & $-$0.61 & 0.96 & $-$0.62 & 0.78 & $-$0.06 & 0.98 & 0.33 & 0.99 \\
    & Bigram  & 0.94 & 0.92 & $-$0.28 & 0.95 & $-$0.40 & 0.78 & 0.16 & 0.98 & 0.52 & 1.00 \\
    & Trigram & 0.98 & 0.92 & 0.27 & 0.94 & $-$0.06 & 0.78 & 0.35 & 0.98 & 0.66 & 1.00 \\
    & $N$     & 40 & 23 & 40 & 26 & 40 & 26 & 40 & 29 & 40 & 28 \\
\addlinespace
GD  & Unigram & $-$0.46 & 0.94 & $-$0.16 & 0.95 & 0.00 & 0.94 & 0.44 & 0.99 & 0.68 & 1.00 \\
    & Bigram  & 0.98 & 0.96 & 0.19 & 0.93 & 0.25 & 0.93 & 0.61 & 0.99 & 0.82 & 0.99 \\
    & Trigram & 0.99 & 0.95 & 0.66 & 0.91 & 0.54 & 0.93 & 0.74 & 0.98 & 0.90 & 0.99 \\
    & $N$     & 40 & 25 & 40 & 27 & 40 & 27 & 40 & 30 & 40 & 28 \\
\addlinespace
TRT & Unigram & $-$0.62 & 0.93 & $-$0.58 & 0.95 & $-$0.44 & 0.96 & $-$0.03 & 0.98 & 0.38 & 0.99 \\
    & Bigram  & 0.94 & 0.95 & $-$0.26 & 0.94 & $-$0.19 & 0.96 & 0.18 & 0.97 & 0.57 & 0.99 \\
    & Trigram & 0.98 & 0.95 & 0.30 & 0.92 & 0.15 & 0.96 & 0.36 & 0.97 & 0.70 & 1.00 \\
    & $N$     & 40 & 24 & 40 & 27 & 40 & 26 & 40 & 29 & 40 & 28 \\
\bottomrule
\end{tabular}
\caption{\blue{Pearson $r$ between $n$-gram-likeness and $\Delta llh$ within each model size and corpus. Columns show all 40 checkpoints (all) or checkpoints from the measure-specific $\Delta llh$ peak through the final checkpoint, including the peak (post). $N$ gives the number of checkpoints for each measure and interval; it is shared across the three $n$-gram orders. Tied peaks use the earlier checkpoint. Checkpoints are dependent, and very small post-peak subsets can yield extreme correlations.}}
\label{tab:ngram-dll-r}
\end{table*}


\end{document}